\documentclass[10pt,twocolumn,letterpaper]{article}

\usepackage[pagenumbers]{cvpr} 

\definecolor{cvprblue}{rgb}{0.21,0.49,0.74}
\usepackage[pagebackref,breaklinks,colorlinks,allcolors=cvprblue]{hyperref}

\def\paperID{9858} 
\def\confName{CVPR}
\def\confYear{2025}

\usepackage{algorithm}
\usepackage{algorithmic}
\usepackage{amsfonts}
\usepackage{xspace}
\usepackage{breqn}
\usepackage{hyperref}
\usepackage{subcaption}
\usepackage{multirow}
\usepackage{enumitem}
\usepackage{xcolor,soul}
\usepackage{graphicx}
\usepackage{subcaption}

\newcommand{\names}{Ges3ViG\xspace}
\newcommand{\namess}{Ges3ViG }

\newcommand{\dataset}{ImputeRefer\xspace}

\newcommand{\aug}{Imputer\xspace}
\newcommand{\augns}{Imputer}

\newcommand{\hey}[1]{\textcolor{red}{\textbf{[{#1}]}}}

\newcommand{\dw}[1]{\colorbox{green}{\parbox{0.5\textwidth}{\bf DW: #1}}}

\newcommand{\noind}[0]{\vspace{5 pt} \noindent}
\newcommand{\noindpar}[1]{\noind {\bf #1}}

\title{\LARGE \bf
\names: Incorporating Pointing Gestures into Language-Based 3D Visual Grounding for Embodied Reference Understanding
}

\author{Atharv Mahesh Mane\textsuperscript{$\star$}\\
\small Stony Brook University, USA\\
\small BITS Pilani Goa campus, India\\
{\tt\footnotesize atharv.mane@stonybrook.edu}
\and
Dulanga Weerakoon\textsuperscript{$\star$}\\
\small Singapore-MIT Alliance for Research \\ \small and Technology Centre, Singapore\\
{\tt\footnotesize dulanga.weerakoon@smart.mit.edu}
\and
Vigneshwaran Subbaraju\\
\small IHPC, Agency for Science, Technology\\ \small and Research (A*STAR), Singapore\\
{\tt\footnotesize vsubbaraju@ihpc.a-star.edu.sg}
\and
Sougata Sen\\
\small BITS Pilani Goa campus, India\\
\small APPCAIR, India\\
{\tt\footnotesize sougatas@goa.bits-pilani.ac.in}
\and
Sanjay E. Sarma\\
\small Massachusetts Institute of Technology, USA\\
{\tt\footnotesize sesarma@mit.edu}
\and
Archan Misra\\
\small Singapore Management University, \\\small Singapore\\
{\tt\footnotesize archanm@smu.edu.sg}
}

\begin{document}
\maketitle
\def\thefootnote{$\bigstar$}{\footnotetext{These authors contributed equally to this work}}

\def\thefootnote{$$}\footnotetext{© 2025 IEEE.  Personal use of this material is permitted.  Permission from IEEE must be obtained for all other uses, in any current or future media, including reprinting/republishing this material for advertising or promotional purposes, creating new collective works, for resale or redistribution to servers or lists, or reuse of any copyrighted component of this work in other works.}
\begin{abstract}
3-Dimensional Embodied Reference Understanding (3D-ERU) 
combines a language description and an accompanying pointing gesture to identify the most relevant target object in a 3D scene. 
Although prior work has explored pure language-based 3D grounding, there has been limited exploration of 3D-ERU, which also incorporates human pointing gestures. 
To address this gap, we introduce a data augmentation framework-- \textbf{\aug}, and use it to curate a new benchmark dataset-- \textbf{\dataset} for 3D-ERU, by incorporating human pointing gestures into existing 3D scene datasets that only contain language instructions. 
We also propose \textbf{\names}, a novel model for 3D-ERU that achieves $\sim$30\% improvement in accuracy as compared to other 3D-ERU models and $\sim$9\% compared to other purely language-based 3D grounding models. Our code and dataset are available at \url{https://github.com/AtharvMane/Ges3ViG}.
\vspace{-0.25in}
\end{abstract}


  
\section{Introduction}
\label{sec:intro}
\begin{figure}
    \centering
    \includegraphics[width=1.0\linewidth,height=1.1in]{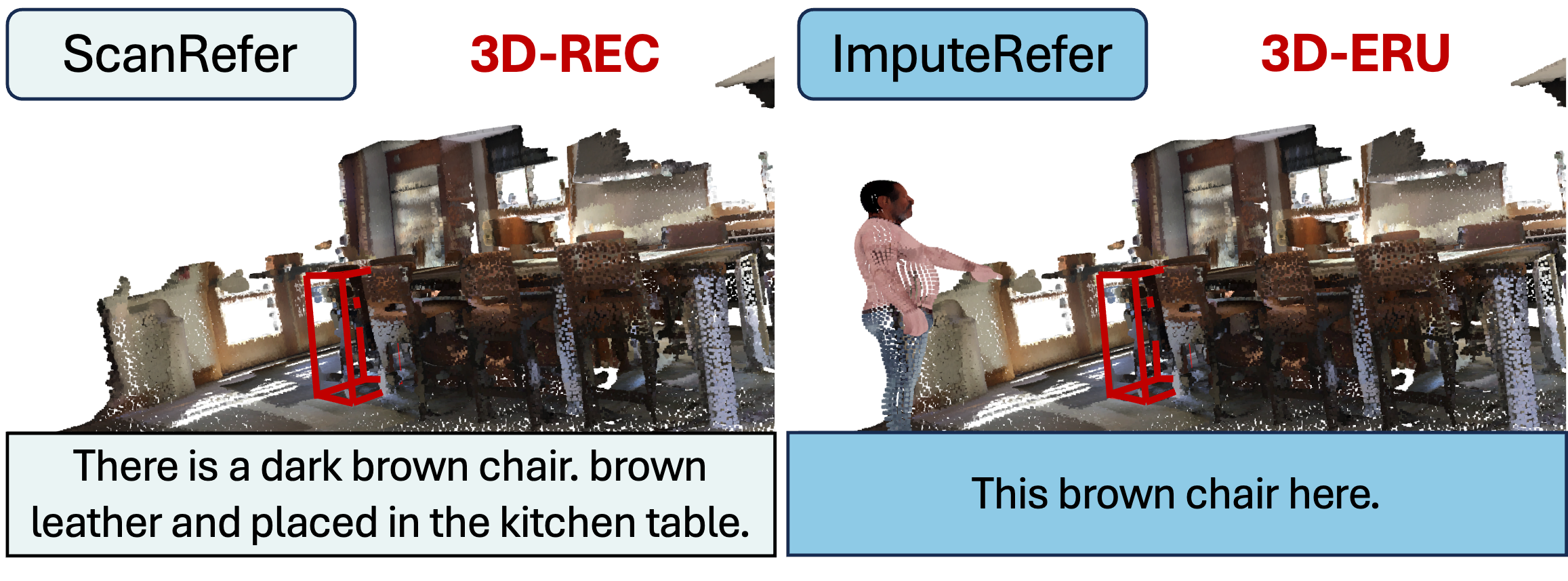}
    \caption{
    3D grounding of referring expressions with pointing.
    \label{fig:introfig}}
    \vspace{-0.25in}
\end{figure}

\renewcommand*{\thefootnote}{\arabic{footnote}}
\setcounter{footnote}{0}
\textit{Referring Expression Comprehension} (REC) is a fundamental vision-language task that involves identifying a target object that is referred to by an instruction, as illustrated in Figure~\ref{fig:introfig}. 
\textit{Grounding} such referring expressions in 3-dimensional space is a critical perceptual capability for robots and intelligent situated agents. Although several studies have explored grounding purely linguistic referring expressions in 3D scenes without accompanying pointing gestures~\cite{chen2020scanrefer, m3drefclip}, pointing is an integral part of human communication and plays a vital role in situated and embodied interactions.
However, the problem of \textit{3D Embodied Reference Understanding (3D-ERU)}~-- which includes natural human pointing gestures along with natural language references for 3D visual grounding remains relatively underexplored. 
This task involves (a)~determination of the human's position and the pointed direction in 3D space, and (b)~identification of an object that best fits the verbal instruction \textit{and} the pointed location.  

To obtain a large workable dataset for 3D-ERU, prior work such as ScanERU~\cite{Lu2024ScanERU}, explored an augmentation approach by inserting a human avatar into the 3D scenes found in a pre-existing 3D-REC dataset called ScanRefer~\cite{chen2020scanrefer}, such that the avatar is pointing to the target object that is being referred by the language instruction. 
However, the following issues remain to be tackled in both data curation and model development for 3D-ERU:
 \begin{itemize}[leftmargin=*]
    \item \textit{Manual positioning of human avatar}: To address the complexities of positioning the human avatar in available and appropriate vacant spaces within the 3D environment, ScanERU employed crowd-sourced human subjects to manually insert the avatar into the 3D scenes provided by ScanRefer. However, our inspection,\footnote{ScanERU is no longer publicly available. We could obtain only a partial test set after contacting the authors.} revealed that the human avatars are placed very close to the target object, thus making the language instruction completely redundant. 
    This is problematic as the models developed using such a dataset will be biased towards learning the pointing gesture alone, disregarding the language references. 
    The scalability of the manual placement approach is also questionable, and an automated method may be preferable for collecting large-scale data. 
    \item \textit{Unrealistic language descriptions}: ScanERU retained the original language descriptions from ScanRefer even after incorporating pointing gestures. This is not realistic as prior work such as \cite{COSM2IC} has shown that the choice of words and verbosity of the language instructions will be different when a pointing gesture is used (see Figure~\ref{fig:introfig}).
    \item \textit{Limited computational model}: The baseline model provided by ScanERU assumes that the ground truth of the human's position in the scene is known and does not localize the human. However, localizing the human is an integral part of 3D-ERU, and hence, a holistic model for 3D-ERU must integrate human localization into its pipeline. We show that effective learning of human position improves the performance of 3D-ERU models.  
 \end{itemize}
To address these issues, we propose a novel automated data augmentation approach along with a revamped instruction dataset for 3D-ERU.
We then develop a new model that incorporates human localization into 3D grounding.
The following are the main contributions of this work:
 \begin{itemize}[leftmargin=*]
    \item We introduce \textbf{\augns}, an automated framework for augmenting existing 3D-REC datasets (e.g., ScanRefer) for 3D-ERU tasks by inserting human avatars into suitable spaces in the 3D scenes with precise control over their positioning, such that they point towards the target. 
    \item We introduce the \textbf{\dataset} dataset for the 3D-ERU task, which requires more complex reasoning to identify target objects than ScanERU due to a larger average distance between the human avatar and the target object. Also, the language instructions in \dataset are re-generated by a VLM that is prompted by providing the context that there is a human pointing at the target. 
    \item We develop \textbf{\names}, a unified model for 3D-ERU that performs both human localization and reference understanding. It implements a new combined loss function to simultaneously learn human-localization and instruction grounding. It also uses a novel multi-stage fusion mechanism to effectively integrate the pointing gesture and the language text to achieve a significant $\sim$29.46\% increase in grounding accuracy compared to ScanERU.
     
\end{itemize}
Overall, we believe that \dataset and \namess help to significantly advance work on 3D-ERU.

\section{Related Work}
\label{sec:related}
Referring Expression Comprehension (REC) on 2D images has been well studied, starting from ReferIt~\cite{kazemzadeh2014referitgame}, and several other works that followed~\cite{mao2016generation, rohrbach2016grounding, yu2018mattnet}. 
Recently, transformer-based architectures~\cite{deng2021transvg} and Vision-Language Models (VLMs)~\cite{wang2023cogvlm} have emerged as promising approaches for 2D-REC. 
The incorporation of pointing gestures into 2D-REC has also been studied by works such as M2Gestic \cite{weerakoon2020gesture}, YouRefIt~\cite{chen2021yourefit} and COSM2IC~\cite{COSM2IC}. However, these 2D RGB image-based approaches are not directly applicable to 3D point cloud data, due to differences in data representation and the significant increase in the search space. 
For 3D-REC, pointcloud-based datasets such as ScanRefer~\cite{chen2020scanrefer} and ReferIt3D~\cite{achlioptas2020referit3d} were developed, with scenes obtained from the ScanNet dataset~\cite{dai2017scannet}. 
Among them, ScanRefer stands out as a large-scale 3D-REC dataset featuring human-annotated language descriptions, comprising 46,055 descriptions of 11,046 different target objects. 
Existing work on 3D-REC encompasses graph-based approaches~\cite{huang2021text, feng2021free}, neuro-symbolic approaches~\cite{hsu2023ns3d}, techniques using multi-view images and 2D semantics~\cite{huang2022multi, yang2021sat, bakr2022look}, and unified models that address both dense captioning and 3D-REC~\cite{chen2022d, cai20223djcg, chen2023unit3d}. 
M3DRefCLIP~\cite{m3drefclip}, currently, the state-of-the-art in 3D-REC, introduces a CLIP model-based approach for extracting visual features~\cite{radford2021learning}. 
However, these works do not consider pointing gestures, which humans naturally use when referring to objects in a scene.

The problem of 3D-ERU, which considers the incorporation of pointing gestures for comprehending referring expressions in a 3D scene remains under-explored compared to 3D-REC. 
ScanERU~\cite{Lu2024ScanERU} proposed inserting a human avatar into an existing 3D-REC dataset. To date, this stands as the sole dataset for 3D-ERU. However, ScanERU's approach for dataset creation suffers from drawbacks such as (a) laborious manual placement of human avatar, (b) positioning the avatar too close to the target, (c) language expressions that are generated without consideration of the pointing gesture, and (d) implicitly assuming that the 3D ground truth location of the human avatar is externally available. These drawbacks hinder both the scalability of the dataset and the performance of the models. 

In contrast to ScanERU, our work adopts an automated approach to insert the human avatar in the 3D scenes. Our proposed dataset consists of more challenging and realistic instructions as the human avatar is placed further away from the target (leading to larger pointing errors) and the language instructions are re-generated using a VLM that also considers the human pointing at the target object. Furthermore, our model includes human localization into the pipeline for 3D-ERU. 
\section{\aug Framework}
\label{sec:method}
We next describe our augmentation framework --~\aug -- to automatically generate a 3D-ERU dataset. 
Similar to ScanERU~\cite{Lu2024ScanERU}, we  use the 3D scenes obtained from ScanNetV2~\cite{dai2017scannet} and insert a human avatar into it.
\aug consists of two parts: 
    (a)~Pointing gesture augmentation, where we propose a simple automated approach to determine the possible locations to \emph{`impute'} the pointing human avatar and 
    (b)~Language description generation, where we employ a Generative model-- Gemini \cite{gemini}, to augment the existing verbal description to incorporate the pointing gesture. 
We next describe each of these components. 
\subsection{Pointing Gesture Augmentation with \aug \label{sec:pointing-ges-gen}}
\label{subsec:imputing}

\begin{figure}
    \centering
    \includegraphics[width=1.0\linewidth,height=1.7in]{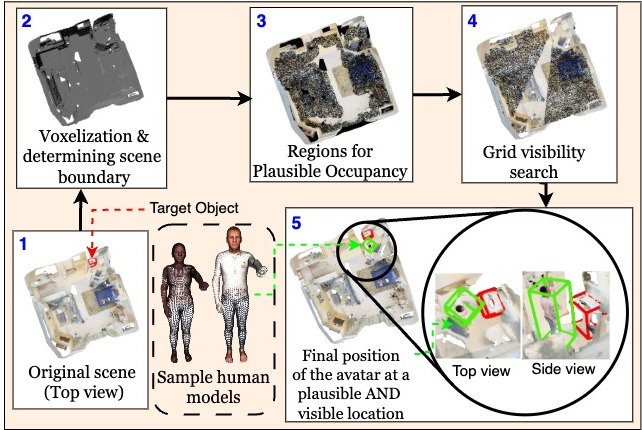}
    \caption{\aug pointing gesture generation}
    \label{fig:imputer_fig}
    \vspace{-0.25in}
\end{figure}

Figure \ref{fig:imputer_fig} illustrates the intermediate steps involved in the pointing gesture augmentation with \aug framework. Before we delve into the computational process involved in inserting a human avatar, we note the following assumptions made by \aug about the 3D scene: (a)~all the 3D scenes are oriented such that the floor's normal is aligned with the Z-axis, (b)~the floor is level (there are no artifacts like stairs) for a major part of the scene, and (c)~the human avatar mesh is pre-aligned with the `floor' of the scene. If the mesh is not pre-oriented correctly, one can use software such as CloudCompare~\cite{cloudcompare} to adjust it. This process only needs to be performed once per avatar, as the same mesh can be reused across all scenes. 

\noindpar{Acquiring realistic human avatars:} Augmenting the scene with a pointing gesture requires generating a human point cloud that can be augmented into the 3D scene such that the generated human point cloud performs a pointing gesture towards the target object. 
We begin this step by extracting the base human meshes for male and female subjects separately from the SMPL-X dataset \cite{SMPL-X}. 
Next, we utilize the Blender package~\cite{Blender} to adjust various model parameters, such as joint poses, height, weight, and gender, to introduce diversity across the generated human point clouds. To add color textures, we use samples from SMPLite-X~\cite{smplitex}, creating a total of 12 textured human models, evenly distributed between genders with six models each.




The workflow of the \aug framework is as follows. We first voxelize the scene and determine its boundaries. Then, we search for a set of vacant spaces for placing the human avatar so that there is a direct line of sight available for pointing at the target. 
The avatar is then placed in a suitable vacant space. 
Finally, a generative model is prompted to generate a language instruction for referring to the target object. These steps are explained in detail below.
    
\noindpar{Voxelization of the scene:} 
We voxelize the scene to discretize the search space, which allows for efficient querying of point occupancy. 
Checking occupancy in a voxel grid is substantially faster compared to meshes or point clouds, and voxel grids allow interaction with `empty space'. 
These empty spaces are crucial, as they define the potential locations where the human avatar can be placed.
We create voxel grid $V_1(\mathbb{N}^3\rightarrow \{0,1 \} )$, which represents the original scene with the target object, and voxel grid $V_2(\mathbb{N}^3\rightarrow \{0,1 \} )$ which represents the scene with the target object removed. Each voxel $V_i(x,y,z)$ in $V_1$ and $V_2$ is assigned a value of `1' if it is occupied by an object or `0' if it is unoccupied.
We also voxelized the human avatar mesh separately to get a 3D grid (similar dimension as $V_1$ and $V_2$) given by $H(\mathbb{N}^3\rightarrow \{0,1 \} )$. A voxel $H(x,y,z)$ is assigned a value of `1' only if it is occupied by the human or to `0' otherwise. The height of the human in the voxel grid is denoted by $h_{hv}$.

\noindpar{Determining scene boundaries: } In this step, we calculate the scene bounds to ensure that the avatar is placed within the scene. This is done by identifying the the largest contour on the planar voxel grid created by projecting the voxel grid $V_1$ onto the $XY$-plane. We then create a binary voxel grid $B(\mathbb{N}^3 \rightarrow {1,0})$, with the same dimensions as $V_1$. A voxel $B(x,y,z)$ is assigned a value of `1' if its $x$ and $y$ coordinates lie within the contour boundary, or `0' otherwise.


\noindpar{Finding regions for plausible occupancy:} Next, we identify all regions within the scene that can accommodate a human mesh. 
To ensure that the feet of the human avatar is placed appropriately, i.e., just touching the floor of the scene, we must determine the appropriate co-ordinates of the human feet $(x_{foot}, y_{foot}, z_{foot})$ and a plausible representative z-coordinate for the floor level within the voxel grid. 
We define the `foot' as the lowest point in the human avatar mesh after converting it into a voxel scale. 
Since there are multiple voxels corresponding to `floor', we adopt the following approach to estimate the vertical displacement of the floor (denoted as $h_{flr}$) with respect to the origin of the scene. 
The 3D scenes in the ScanRefer dataset originate from ScanNetV2~\cite{dai2017scannet}. 
Using the segmentation ground truth from ScanNetV2, we can extract a list of vertices labeled as `floor'.  
We calculate $h_{flr}$  by taking the minimum of the following two values: 
    (1) the average height of all floor vertices plus an offset $C_1 = 0.04$m, and 
    (2) the 85\textsuperscript{th} percentile of the height distribution of `floor' points. Here, $C_1$ and 85\textsuperscript{th} percentile are chosen empirically.
Next, we convert this estimated height to a voxel scale denoted as $h_{fv}$. The final floor height estimated is given by $\hat{h}_{fv} = max(C_2, h_{fv})$, where $C_2=4$. $C_2$ is chosen to account for potential artifacts, such as floor misalignment, that could cause it to extend into voxels at heights above zero. 
We allow for a small tolerance by permitting the human model to be positioned up to 4 voxels ($\sim$10 cm in world coordinates) above the floor. In practice, this offset is generally less than 5 cm.


We then adopt a sliding volume approach to identify sufficiently large vacant spaces that can fit the human avatar. 
The size of the sliding volume is chosen to be equal to the size of the human avatar voxel grid. 
We then use the human voxel grid as a mask to check if there is sufficient space to fit the human avatar at a given point $(i,j)$  on the floor. 
We consider the human avatar volume $H$ comprising of voxels whose co-ordinates range from  $(i,j,\hat{h}_{fv})$ to $(i+x_{max}, j+y_{max}, \hat{h}_{fv}+h_{hv})$. We see that the entire volume falls into a clear space of the scene if it satisfies,  
    \begin{dmath}\label{eq: occ_points}   \sum_{x=0}^{x_{max}}\sum_{y=0}^{y_{max}}\sum_{z=0}^{h_{hv}}H(x,y,z)= \sum_{x=0}^{x_{max}} \sum_{y=0}^{y_{max}}\sum_{z=0}^{h_{hv}}{H(x,y,z) \times V_1(i+x,j+y, \hat{h}_{fv}+z)}
    \end{dmath}
In practice, we observed that there could be some cases of improper human positioning even after the above condition is satisfied (especially after rotation of the avatar). 
To avoid such scenarios, we increase the border widths by 25 cm (10 voxels), which is approximately half the average shoulder-to-shoulder width of a human. 
We store the co-ordinates $(i,j,\hat{h}_{fv}+h_{hv})$ in the set $O_{no\_collide}$ if all the voxels in the volume placed at $(i,j)$ on the floor satisfy Equation~\ref{eq: occ_points} after accounting for the additional allowance. 



\noindpar{Grid visibility search: } In this step, we aim to identify all the voxels that are accessible via a direct line-of-sight (a spatially contiguous set of empty voxels) from the center of the object. These voxels would serve as plausible locations from where the human avatar could point at the object. We define the center of the 3D bounding box of the target object as the origin of the line of sight. We use the voxel grid $V_2$, where the target object is removed for visibility search. This is done so that the target object's own boundary surface does not interfere with line-of-sight calculation. 

Prior studies (e.g., \cite{gridwalk}) have used a path-counting algorithm in 2D space to calculate visibility scores for grid cells. We extend this approach to 3D environments to compute the regions with line-of-sight. Given $V_2$, we compute a visibility score grid, $S(\mathbb{N}^3 \rightarrow \mathbb{R})$, where $S$ is defined as:
\begin{multline}   
    S(x,y,z) = (\frac{x \times S(x-1,y,z)}{x+y+z} + \frac{y \times S(x,y-1,z)}{x+y+z} \\ + \frac{z \times S(x,y,z-1)}{x+y+z} ) \times V_2(x,y,z)
    \vspace{-0.2in}
\end{multline}
We then obtain $S'(\mathbb{N}^3 \rightarrow \{1,0\})$ a thresholded version of $S$, where $S'(x,y,z)$ is assigned a value of `1' if its visibility score is more than 0.33, or set to `0' otherwise. The value 0.33 was chosen using a rough rule of thumb -- $\frac{1}{No. of dimensions}$. In \cite{gridwalk}, a threshold of $0.5$ was used to calculate visibility in 2D grids. 
We define the set of all the coordinates where $S'=1$, as the \emph{region of visibility}, a good approximation for all the visible regions.

\noindpar{Final positioning of the human avatar: } 
A correctly \emph{imputed} human avatar, with its feet at (x,y,z) and pointing at the referred object, must satisfy the following conditions: 
    (a)~the human avatar is placed within the bounds of the scene, i.e, $B(x,y,z)=1$, 
    (b)~the avatar does not occupy any already-occupied space in the scene, i.e, if $(x,y,z) \in O_{no\_collide}$, and 
    (c) there is a direct unblocked line between the human avatar's gesture and the object of interest in the scene, i.e., if $S'(x,y,z)=1$.
After identifying all such \emph{feasible} points, we randomly choose 5 of these points to create diverse, feasible positions of the human avatar. 
For each of these points, we then use the gesturing shoulder of the human avatar as a pivot of rotation to point towards the object of interest. 
In natural scenarios, a human may have some directional error while pointing towards a target object. To simulate this, we introduce a rotational jitter sampled uniformly within a range of $N_{jitter} = 9^{\circ}$ to the correct pointing direction. 
We then calculate simple Euclidean transforms to move the human to the desired coordinate and then rotate the human in place. To choose from appropriately pointing meshes, we also store the angle $\angle A_{pointing}$ between the X-Y plane and the line joining the shoulder point to the desired pointing direction. This completes the process of augmenting the 3D scene with an \emph{appropriately-posed} human avatar as shown in Figure~\ref{fig:imputer_fig}, where the avatar (green box) points towards the target object (red box). 

\begin{figure}[t]
    \centering
        \includegraphics[width=1.0\linewidth,height=1.5in]{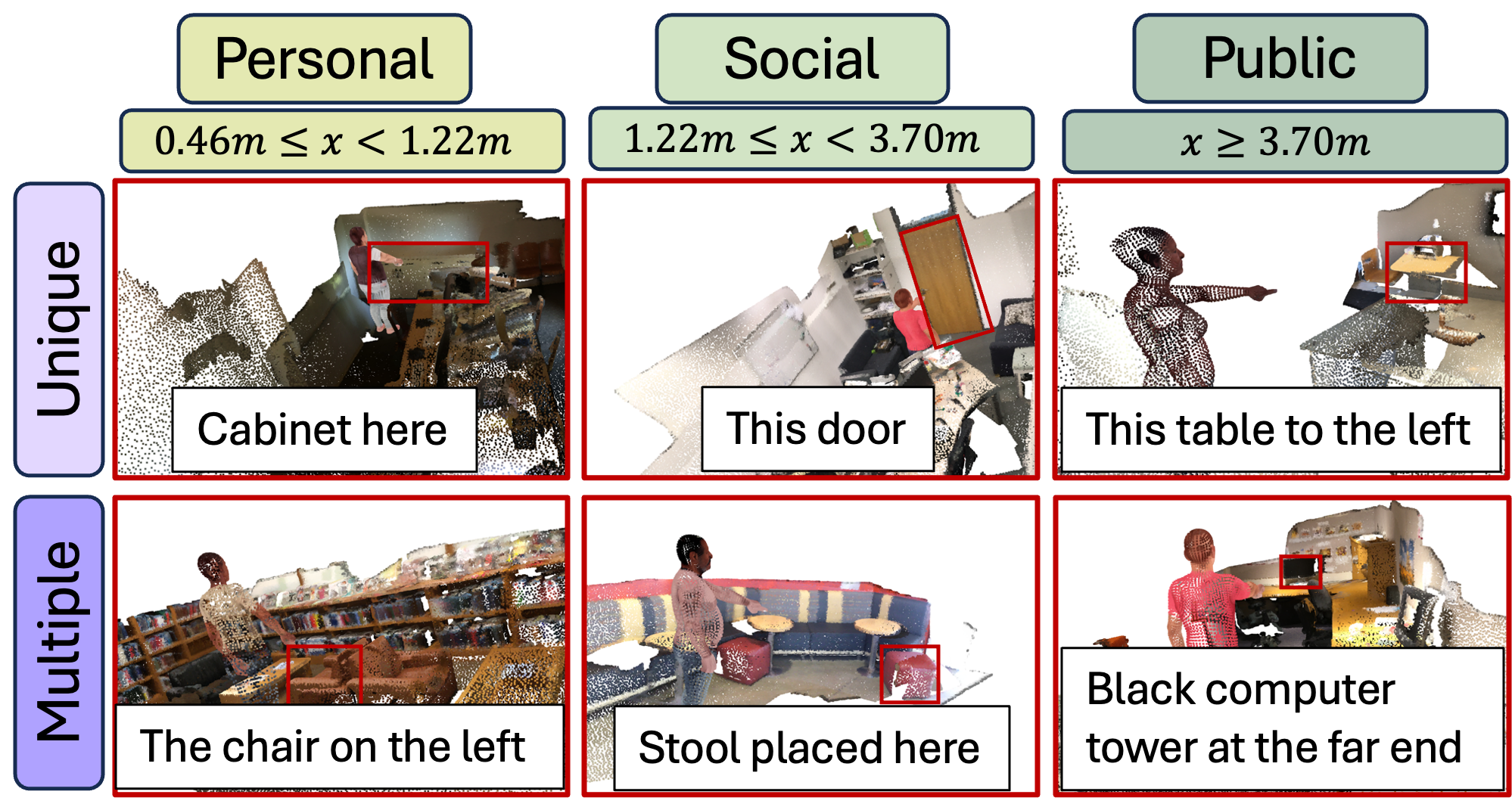}
        \caption{\dataset with `unique' and `multiple' object scenes. The imputed human is placed at different distances from the target.}  
        \label{fig:impute_samples}        
        \vspace{-0.2in}
\end{figure}

\subsection{Referring Expression (Re)Generation:} 
\label{subsec:gemini}
We use Gemini~\cite{gemini} to augment the referring expression. 
The input prompt to Gemini for generating the referring expression included a brief paragraph describing the scenario: 
\textit{a person pointing at the target object needs to generate an expression that uniquely identifies that object}. To ensure diversity in the generated expressions, for each data point in the ScanRefer \cite{chen2020scanrefer} dataset, we prompt Gemini to generate 3 different referring expressions and choose one of them randomly to generate our new dataset \dataset for 3D-ERU, some samples of which are shown in Figure~\ref{fig:impute_samples}. 
Next, we describe the development of our \names model using this dataset.


\section{\texorpdfstring{\names} \ : Gesture-enhanced 3D Visual Grounding}
\label{sec:model}

\begin{figure}
    \centering
    \includegraphics[width=0.99\linewidth]{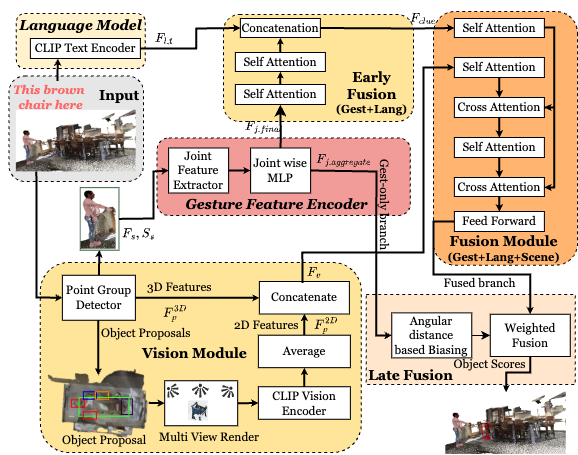}
    \caption{Proposed architecture of \namess}
    \label{fig:architecture}
    \vspace{-0.25in}
\end{figure}

We now present \names, a new gesture-enhanced, unified feed-forward, 3D visual grounding DNN model that integrates human localization, language understanding, and gestural reasoning to identify the 3D bounding box of the target object. To effectively combine gestural and linguistic cues, we utilize an innovative multi-stage fusion strategy that employs both early and late fusion within the same DNN. We now highlight the key components of the \names model, visually illustrated in Figure~\ref{fig:architecture}.

\subsection{\textbf{Vision Module}\label{sec:vismod}} We first encode the point cloud features using a PointGroup detector \cite{pointgroup}. PointGroup, being a standard object detector for 3D point clouds $P\in \mathbb{R}^{N\times C}$, returns both the visual features and all the objects in the scene. 
Here, $N$ is the number of points in the point cloud and $C$ is the number of channels per point, including the 3D coordinate, the per-point surface normal vector and the color of the point.\\
We use the following outputs from the PointGroup Detector for further processing: 
(1)~Semantic Features ($F_s \in \mathbb{R}^{N\times32}$), (2)~Semantic Scores ($S_s \in \mathbb{R}^{N\times N_c}$), where $N_c$ is the number of classes in the dataset, (3)~Proposal Features ($F^{3D}_p \in \mathbb{R}^{M\times 32}$), where $M$ is the number of proposals, and (4)~Proposal Scores ($S_p \in \mathbb{R}^{M}$): These are essentially the instance/objectness scores for each proposal. 

We use $F_s$ and $S_s$ for Gesture Inference (details in Section~\ref{sec:GesEnc}), while $S_p$ is used to find proposals and $F^{3D}_p$ is passed ahead to be used as 3D visual features.
For each proposal $P$, we render the object from 3 different views as images (similar to M3DRef-CLIP~\cite{m3drefclip}). The images are  passed into a CLIP~\cite{clip} image encoder that generates 2D features ($F^{2D}_p \in \mathbb{R}^{M \times 128}$) for each proposal. $F^{3D}_p$ and $F^{2D}_p$ are concatenated into joint visual features and passed through a 1D convolution to obtain $F_{v} \in \mathbb{R}^{M \times 128}$.


\subsection{\textbf{Language Feature Encoder }} For language features, we use CLIP to get sentence-level and token-level feature embeddings. We define our sentence features as $F_{l,s} \in \mathbb{R}^{128}$ and word features as $F_{l,t} \in \mathbb{R}^{N_{token}\times 128}$, for an $N_{token}$ token long sequence.

\subsection{\textbf{Gesture Feature Encoder}\label{sec:GesEnc}} 


To enable gestural reasoning for human segments identified from region proposals generated by the PointGroup detector, this module uses the following components.

    \noindpar{Joint Feature Extractor: } This component computes the joint-coordinates of the pointing human and extracts joint-based features. The semantic features ($F_s$) computed in the vision module are used as input to this component. For a human segment $P_{hum} \subset P$, obtained from the semantic segmentation predictions of the Vision Module and comprising $N_{hum}$ points, we extract a corresponding subset of semantic features, $F_{s,hum} \subset F_s$, consisting of $N_{hum}$ vectors. Given an anticipated number of human joints, $N_j \in \mathbb{N}$, the semantic feature subset $F_{s,hum}$ is processed through a point-wise MLP inspired by PointNet \cite{Pointnet}, resulting in an output $W_{hum} \in \mathbb{R}^{N_{hum} \times N_j}$. Inspired from Point2Skeleton \cite{Point2Skeleton}, the joint coordinates $J \in \mathbb{R}^{N_j \times 3}$ are then computed from $W_{hum}$ as 
       $ J = Softmax(W_{hum}^T) \times P_{hum}$
    and the initial joint-based features $F_{j, init} \in \mathbb{R}^{N_j,32}$ are obtained as
    \begin{equation}
        F_{j,init} = Softmax(W_{hum}^T) \times F_{s,hum}
    \end{equation}
    
    \noindpar{Joint-Wise MLP: } We concatenate the location of each joint with its corresponding feature to form $F_{j,init}$. This concatenated feature is processed by Joint-wise MLP, which consists of $N_{layers}$ layers. In this configuration, the MLP processes each feature vector independently for each joint. Let $M_i$ represent the $i$-th layer of the MLP, and let $P_{j,i}$ denote its output for the $j$\textsuperscript{th} joint. This module's output is:
    \begin{multline}
        F_{j,final} = P_{j,N_{layers}}~; ~where ~P_{j,i} = M_{i}(P_{j,i-1}) \\ and ~P_{j,1} = M_{1}(F_{j,init})
    \end{multline}
    \begin{equation}
        F_{j,aggregate} = concatenate_{i=0}^{i=N_{layers}}{\sum_{j=0}^{N_{j}}{P_{j,i}}}
    \end{equation}
    Here $N_j$ is the number of joints specified in SMPL.

We then adopt a multi-stage fusion approach that combines early and late fusion techniques to effectively fuse the extracted gesture features with language features. We first perform an early fusion of the `reference' modalities by concatenating gesture and language instruction features. The concatenated features are fed into the `reference-scene' fusion module, which fuses them with the 3D scene/object features obtained from the vision module to produce prediction scores for the potential target objects. Finally, a late weighted fusion biases the scores towards objects that are within a certain angular radius of the pointing hand. 

\subsection{\textbf{Early Fusion of Referential Modalities}}

In the proposed early fusion module, we utilize $F_{j, final} \in \mathbb{R}^{N_j \times 128}$, which is passed through two self-attention layers, after which they are concatenated with $F_{l, t}$ to form $F_{clue} \in \mathbb{R}^{(N_{token} + N_j) \times 128}$. Since both gesture and language modalities provide information about the same target object, we treat them similarly and perform a simple concatenation-based fusion for the two modalities.


\subsection{\textbf{Fusion of Referential and Scene Features}}
We employ the same transformer-based approach for the `reference-scene' fusion module as used in M3DRef-CLIP~\cite{m3drefclip} to integrate object features with the referential embeddings $F_{clue}$, generating confidence scores ($S_{conf} \in \mathbb{R}^{M}$) for each proposal. This module consists of two self-attention and two cross-attention blocks, where the object features sequentially pass through the self-attention and cross-attention blocks. In the cross-attention blocks, $F_{clue}$ is used as the key-value pairs.

\subsection{\textbf{Late Fusion of Gesture}}

The reference-scene fusion module's output  provides confidence scores for each object proposal, allowing to identify the target object based on the highest score. 
To further refine this, we introduce an additional late fusion approach that explicitly biases confidence scores to favor objects that align with the pointed direction. 
This corrects some errors from the previous stages that could occur due to the presence of multiple distractor objects, especially in cases where the language instruction was given more importance. 
Specifically, we compute a pointing biasing score,  $S_b \in \mathbb{R}^{M}$, and add it to $S_{conf}$ to guide predictions toward points that are angularly closer to the pointing direction.
To achieve this, we first extract the shoulder and fingertip points represented by $\vec{v_1} \in \mathbb{R}^{3}$ and $\vec{v_2} \in \mathbb{R}^{3}$, respectively. The center coordinates of a predicted bounding box are represented by $\vec{v_3} \in \mathbb{R}^{3}$.
The pointer bias score $S_b$ is defined as follows:
\begin{equation} \label{eqn:Pointerbiasingeqn}
    S_b = \frac{(\vec{v_2}-\vec{v_1})\cdot (\vec{v_3}-\vec{v_1})}{||((\vec{v_2}-\vec{v_1}))||\cdot||((\vec{v_3}-\vec{v_1}))||}
\end{equation}

We calculate $S_{b,left} \in \mathbb{R}^{M}$ and $S_{b,right} \in \mathbb{R}^{M}$ using the equation \ref{eqn:Pointerbiasingeqn} for both left and right hand respectively.

Intuitively, humans may use either the left or right hand for pointing. To classify the pointing hand used, we employ a classification head that takes $F_{j,aggregate}$ as input to learn $W_{lr_agg} \in \mathbb{R}^{2}$ as follows.
\begin{equation}
    W_{lr\_agg} = Softmax(MLP1(F_{j,aggregate}))
\end{equation}
\par With the calculated $W_{lr\_agg}$, we compute the final biasing score $S_{b,final}$ as follows. 
\begin{equation}
    S_{b,final} = W_{lr\_agg}^0 \times S_{b,left} + W_{lr\_agg}^1 \times S_{b,right}
\end{equation}
With $S_{b,final}$ from the pointing biasing score and $S_{conf}$ from the fusion module, we have two sets of object scores. To obtain the final object scores, we use an additional probabilistic head, denoted as $W_{score_agg} \in \mathbb{R}^{2}$, which adaptively aggregates scores from the two branches as follows: $W_{score\_agg} = SoftMax(MLP2(F_{j,aggregate}))$.
The final object scores are obtained using the following equation, and the target object is identified by the object proposal with the highest $S_{conf,final}$.
\begin{equation}
    S_{conf,final} = W_{score\_agg}^0 \times S_{conf} + W_{score\_agg}^1 \times S_{b,final}
\end{equation}
\subsection{\textbf{Loss Function: }}
Our proposed loss function comprises of the following: \\
\textbf{Object Detection Losses:}
\begin{enumerate}
    \item Cross-entropy loss for supervising per-point semantic class prediction called the Semantic Loss, $L_{semantic}$. 
    \item L1 loss for supervising per-point offset vector towards object centers (useful for instance segmentation), called the Offset Norm Loss, $L_{offset\_norm}$.
    \item A directional loss formed as a mean of minus cosine similarities to constrain the direction of per-point offset vectors called the Offset Direction Loss, $L_{offset\_dir}$.
    \item A binary cross-entropy loss for supervising per-point objectness confidence score called the Score Loss, $L_{score}$.
    \end{enumerate}
\textbf{Reference Understanding Losses:}
\begin{enumerate}[start=5]
    \item Reference Loss, $L_{ref}$, is a multi-class cross-entropy loss to classify object referenced by the referring expressions.
    \item A symmetric contrastive loss, $L_{contrastive}$ between sentence features and the visual features to aid training. 
\end{enumerate}
\textbf{Human Localization Losses:}
\begin{enumerate}[start=7]
\item L2 loss between the predicted and ground truth human joint locations.
    \item A classification loss $L_{lr}$ over $W_{lr\_agg}$ to classify the gesture as left-handed or right-handed.
\end{enumerate}

\subsection{\textbf{Training Details: }} We utilize an iterative training approach where we first train the PointGroup detector and gesture feature encoder for 20 epochs with a learning rate of 0.00005 with cosine annealing, using the Adam optimizer. At this step, loss term $L_{semantic}$, $L_{offset\_norm}$, $L_{offset\_dir}$, $L_{score}$, $L_{hum}$ and $L_{lr}$ are enabled for optimization. Subsequently, we freeze the PointGroup detector and the gesture feature encoder and only enable the fusion module with the loss terms $L_{ref}$ and $L_{contrastive}$ enabled for 20 epochs. This step trains the comprehension module to comprehend verbal instruction.
 \section{Results}
\label{sec:results}
\vspace{-0.1in}We use our new \dataset dataset, to evaluate the performance of various models for the 3D-ERU task. \dataset includes $707$ point-cloud scenes sourced from the original ScanNetv2 dataset, augmented with human pointing gestures and modified referring expressions as described in Sections~\ref{subsec:imputing} and~\ref{subsec:gemini}. Overall, our dataset encompasses $35,581$ samples, with $9,391$ designated as the test dataset. Similar to previous 3D grounding research, we adopt $IoU@0.25$ and $IoU@0.5$ as the evaluation metrics. 
$IoU@0.25$ and $IoU@0.5$ deem a prediction as correct if the Intersection over Union between the ground truth and predicted 3D bounding boxes $\ge0.25$ and $\ge0.5$, respectively.

\subsection{Performance of \names for 3D-ERU}

Table \ref{tab:acc} presents a comparison of the accuracy of \namess against current state-of-the-art models for 3D visual grounding, both with and without gesture support. Notably, ScanERU is the only existing model that incorporates gesture-enhanced 3D visual grounding. From the results, it is evident that \names surpasses ScanERU significantly, achieving a 29.87\% higher overall IoU@0.5 accuracy. In the presence of `multiple' distractors, \names demonstrates even more pronounced improvements of 32.71\% in IoU@0.5 over ScanERU. 

\names achieves significant performance improvements over existing standard 3D visual grounding models that do not support pointing gestures. 
It surpasses M3DRefCLIP, the current state-of-the-art in 3D visual grounding, by 10.88\%, 8.50\%, and 8.93\% for `unique', `multiple', and `overall' categories respectively. 
These cumulative performance gains across both standard 3D visual grounding and gesture-enhanced models highlight the effectiveness of \names in integrating both gestural and linguistic reasoning within a unified architecture. 
\begin{table}[]
 \caption{Performance of \namess Vs baselines on \dataset}
 \label{tab:acc}
 \resizebox{1.05\columnwidth}{!}{
\begin{tabular}{l|ll|ll|ll}
\hline
\multicolumn{1}{c|}{\multirow{2}{*}{\textbf{Model}}} & \multicolumn{2}{c|}{\textbf{unique}}                                                                                                                                & \multicolumn{2}{c|}{\textbf{multiple}}                                                                                                                              & \multicolumn{2}{c}{\textbf{overall}}                                                                                                                               \\ \cline{2-7} 
\multicolumn{1}{c|}{}                                & \multicolumn{1}{c}{\textbf{\begin{tabular}[c]{@{}c@{}}IoU\\ @0.25\end{tabular}}} & \multicolumn{1}{c|}{\textbf{\begin{tabular}[c]{@{}c@{}}IoU\\ @0.5\end{tabular}}} & \multicolumn{1}{c}{\textbf{\begin{tabular}[c]{@{}c@{}}IoU\\ @0.25\end{tabular}}} & \multicolumn{1}{c|}{\textbf{\begin{tabular}[c]{@{}c@{}}IoU\\ @0.5\end{tabular}}} & \multicolumn{1}{c}{\textbf{\begin{tabular}[c]{@{}c@{}}IoU\\ @0.25\end{tabular}}} & \multicolumn{1}{c}{\textbf{\begin{tabular}[c]{@{}c@{}}IoU\\ @0.5\end{tabular}}} \\ \hline
\textbf{without Gestures:}                               &                                                                                  &                                                                                  &                                                                                  &                                                                                  &                                                                                  &                                                                                 \\
3DVG-Transformer \cite{zhao2021_3DVG_Transformer}                                    & 71.56                                                                             & 50.66                                                                            & 31.35                                                                            & 21.54                                                                            & 39.17                                                                            & 27.20                                                                           \\
HAM \cite{chen2022learning}                                                 & 67.10                                                                            & 48.13                                                                            & 25.42                                                                            & 16.04                                                                            & 33.51                                                                            & 22.27                                                                           \\
3DJCG  \cite{cai20223djcg}                                              & 75.93                                                                            & 59.19                                                                            & 40.34                                                                            & 30..61                                                                           & 47.24                                                                            & 36.16                                                                           \\
RefMask-3D  \cite{he2024refmask3d}                                              & 72.45                                                                            & 64.95                                                                            & 25.44                                                                            & 22.24                                                                           & 34.57                                                                            & 30.54                                                                           \\
M3DRefCLIP \cite{m3drefclip}                                          & 77.32                                                                             & 60.15                                                                             & 62.62                                                                             & 47.27                                                                             & 65.53                                                                             & 49.78                                                                            \\ \hline
\textbf{with Gestures:}                                &                                                                                  &                                                                                  &                                                                                  &                                                                                  &                                                                                  &                                                                                 \\
ScanERU \cite{Lu2024ScanERU}                                            & 71.6                                                                             & 52.79                                                                            & 31.91                                                                            & 23.06                                                                            & 39.54                                                                            & 28.84                                                                           \\
\textbf{\names}                                               & \textbf{84.60}                                                                             & \textbf{71.03}                                                                             & \textbf{67.57}                                                                             & \textbf{55.77}                                                                             & \textbf{70.85}                                                                             & \textbf{58.71}                                                                            \\ \hline
\end{tabular}
}
\end{table}

\subsection{Evaluating \aug Augmentation for 3D-ERU}\label{subsec:evalAug}

We evaluate the impact of \aug augmentation on Gesture-enhanced 3D visual grounding in Table~\ref{tab:acc_imputer}. While all results here use the \names model, the first row (ScanRefer) simply integrates the human pointing gestures into the scene while retaining the original pointing-unaware ScanRefer verbal instructions.
We find that a ScanRefer instruction contains a substantially 
higher average word count per language description (19.18 words) compared to \dataset (10.52 words), indicating that ScanRefer likely has many redundant words. However, using the ScanRefer text instructions only results in a slightly better overall IoU@0.5 (59.69\%) than using the referring expressions in \dataset (58.71\%), despite the detailed verbal description. In practice, however, humans generally use short verbal descriptions when using a pointing gesture, justifying our curation of pointing-augmented \dataset dataset. Furthermore, compared to the average human-to-target distance of $\sim$1.24 meters in the shared subset of ScanERU instructions that we obtained, \dataset has an average distance of $\sim$2.31 meters. This approx. 2$\times$ increase in distance increases the complexity of gestural reasoning needed.


        

\begin{table}[t]
\caption{Effect of re-generated instructions in \dataset} \vspace{-0.1in}
\label{tab:acc_imputer}
\centering
 \resizebox{1.05\columnwidth}{!}{
\begin{tabular}{p{25mm}|c|ccc}
\hline
\multirow{2}{*}{\textbf{Text Inst. Source}} & \multirow{2}{*}{\textbf{\begin{tabular}[c]{@{}c@{}}Average \\ No. of words\end{tabular}}} & \multicolumn{3}{c}{\textbf{IoU@0.5 on Ges3ViG}}                                                  \\ \cline{3-5} 
                                  &                                                                                           & \multicolumn{1}{c|}{\textbf{unique}} & \multicolumn{1}{c|}{\textbf{multiple}} & \textbf{overall} \\ \hline
From ScanRefer                   & 19.18                                                                                     & \multicolumn{1}{c|}{71.77}           & \multicolumn{1}{c|}{56.75}             & 59.69            \\ \hline
Generated                      & 10.52                                                                                     & \multicolumn{1}{c|}{71.03}           & \multicolumn{1}{c|}{55.77}             & 58.71            \\ \hline
\end{tabular}
}
\vspace{-0.1in}
\end{table}

\begin{table}[]
 \caption{Ablation studies for \names in \dataset dataset} \vspace{-0.1in}
 \label{tab:ablations}
  \resizebox{1.05\columnwidth}{!}{
\begin{tabular}{l|ll|ll|ll}
\hline
\multicolumn{1}{c|}{\multirow{2}{*}{\textbf{Model}}} & \multicolumn{2}{c|}{\textbf{unique}}                                                                                                                                & \multicolumn{2}{c|}{\textbf{multiple}}                                                                                                                              & \multicolumn{2}{c}{\textbf{overall}}                                                                                                                               \\ \cline{2-7} 
\multicolumn{1}{c|}{}                                & \multicolumn{1}{c}{\textbf{\begin{tabular}[c]{@{}c@{}}IoU\\ @0.25\end{tabular}}} & \multicolumn{1}{c|}{\textbf{\begin{tabular}[c]{@{}c@{}}IoU\\ @0.5\end{tabular}}} & \multicolumn{1}{c}{\textbf{\begin{tabular}[c]{@{}c@{}}IoU\\ @0.25\end{tabular}}} & \multicolumn{1}{c|}{\textbf{\begin{tabular}[c]{@{}c@{}}IoU\\ @0.5\end{tabular}}} & \multicolumn{1}{c}{\textbf{\begin{tabular}[c]{@{}c@{}}IoU\\ @0.25\end{tabular}}} & \multicolumn{1}{c}{\textbf{\begin{tabular}[c]{@{}c@{}}IoU\\ @0.5\end{tabular}}} \\ \hline
\names\textsubscript{w/o Gestures}                                         & 69.26                                                                             & 48.84                                                                             & 52.58                                                                             & 37.32                                                                             & 55.79                                                                             & 39.54                                                                            \\
\names\textsubscript{noHumanLoss}                                         & 68.76                                                                             & 48.57                                                                             & 58.75                                                                             & 42.31                                                                             & 60.68                                                                             & 43.51                                                                            \\
\names\textsubscript{noEF\_onlyLF}                                         & 69.43                                                                             & 49.28                                                                             & 54.81                                                                             & 39.02                                                                             & 57.62                                                                             & 41.00                                                                          \\
\names\textsubscript{onlyEF\_noLF}                                      & 83.71                                                                             & 70.09                                                                             & 66.47                                                                             & 54.92                                                                             & 69.93                                                                             & 58.05                                                                            \\
\names\textsubscript{random\_LF}                                         & 84.0                                                                             & 70.6                                                                             & 66.1                                                                             & 54.6                                                                             & 69.6                                                                             & 57.7                                                                            \\
\names\textsubscript{onlyGest}                                     & 15.29                                                                            & 11.81                                                                            & 12.46                                                                            & 9.80                                                                            & 13.0                                                                            & 10.18                                                                           \\
\names\textsubscript{ConstantLang}                                     & 51.05                                                                            & 43.87                                                                            & 44.89                                                                            & 36.78                                                                            & 46.08                                                                            & 38.15                                                                           \\
\textbf{\names}                                & \textbf{84.60}                                                                             & \textbf{71.03}                                                                             & \textbf{67.57}                                                                             & \textbf{55.77}                                                                             & \textbf{70.85}                                                                             & \textbf{58.71}                                                                            \\ \hline
\end{tabular}
}
\vspace{-0.2in}
\end{table}

\subsection{Ablation Studies}

Table~\ref{tab:ablations} summarizes the results of an ablation study to evaluate the effectiveness of early fusion, late fusion and human localization loss. We consider the following variants: 
\begin {itemize}
\item \textbf{\names\textsubscript{w/o Gestures}} \names without any gestural reasoning and only relying on language-based reasoning.
\item \textbf{\names\textsubscript{onlyEF\_noLF}} integrates the gesture information only once at the early fusion and disables the late fusion of gesture information by assigning a weight of 0 to the gesture branch in late fusion.
\item \textbf{\names\textsubscript{noEF\_onlyLF}} integrates the gesture information only at the late weighted fusion stage and skips the early fusion block. To simulate this variant, we re-train this variant from scratch by directly feeding the language features to the fusion module (by-passing early fusion).
\item \textbf{\names\textsubscript{random\_LF}} uses the full \names pipeline but with random weights in the late fusion.
\item \textbf{\names\textsubscript{onlyGest}} relies solely on the gesture and ignores all language information by assigning a full bias weight of 1 to the gesture branch in weighted late fusion.
\item \textbf{\names\textsubscript{noHumanLoss}} does not include the human localization loss component in the loss calculation. This network is also obtained by re-training from scratch.
\item \textbf{\names\textsubscript{ConstantLang}} All language references set to ``This object'', forcing the model to heavily rely on pointing.
\end{itemize}

From Table~\ref{tab:ablations}, we observe that early fusion is more effective than late fusion for fusing the referential modalities as \names\textsubscript{onlyEF\_noLF} provides significantly high performance than \names\textsubscript{noEF\_onlyLF}. 
But the performance of \names\textsubscript{onlyEF\_noLF} is still slightly lower than \names. 
The performance of \names\textsubscript{random\_LF} is even lower than \names\textsubscript{noEF\_onlyLF}, showing that the weighted late fusion is meaningful as 
it provides a small boost in accuracy as 
compared to early fusion. 
We further note that ignoring the human localization loss causes a significant difference of about 15\% in IoU@0.5 for \names\textsubscript{noHumanLoss} when compared to \names. This underscores the importance of learning to localize the human accurately. \names stands out as the first unified model that performs human localisation and reference understanding. \names\textsubscript{onlyGest} variant that relies solely on gestural scores achieves significantly lower IoU@0.5 of 10.18\% compared to \names (58.71\%). Similarly, \names\textsubscript{ConstantLang} which relies heavily on pointing with the same language reference used across the dataset achieves significantly lower IoU@0.5 of 38.15\%. These findings are supported by prior studies \cite{weerakoon2020gesture,COSM2IC} that showed that language references are important for reference understanding when the target object is further away from the human, and in contrast to ScanERU, where the pointing was done from a much closer distance. This result also shows that Gemini's language instructions in \dataset are meaningful and useful. 

\begin{figure}
    \centering
    \includegraphics[width=0.99\linewidth]{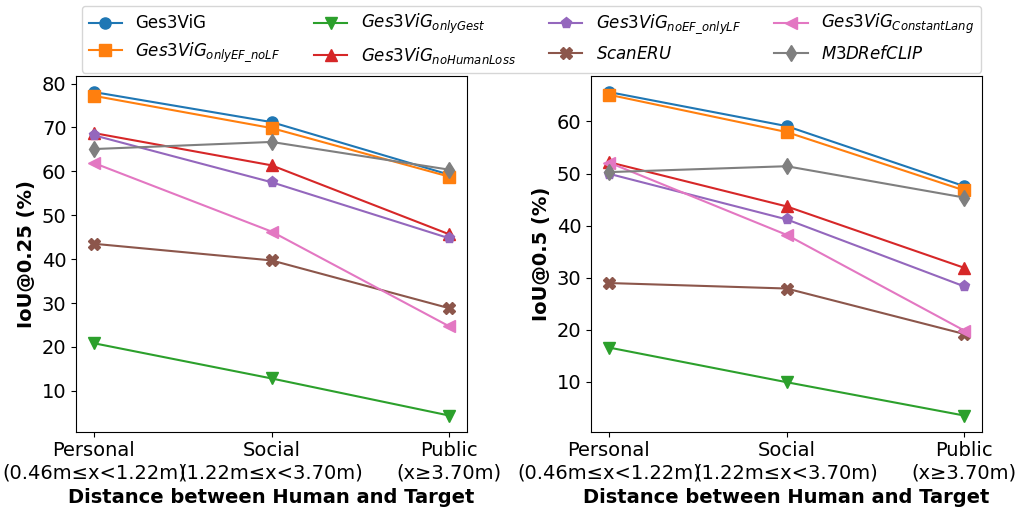}
    \caption{Accuracy at different distance ranges for 3D-ERU}
    \label{fig:dist_vs_acc}
    \vspace{-0.21in}
\end{figure}

Figure \ref{fig:dist_vs_acc} summarizes the effect of distance between the human and the target object on the accuracy of \names model variants and baselines. For this analysis, we divided the test set of \dataset into the following four distance ranges, using Hall's interpersonal distance classification~\cite{hall1963system}--- \textit{Intimate} ($x<0.46m$), \textit{Personal} ($0.46m\leq x < 1.22m$), \textit{Social} ($1.22m\leq x < 3.70m$), and \textit{Public} ($x\geq 3.70m$). The test set of \dataset contained 0 samples for the Intimate distance range, 1391 for Personal, 6909 for Social, and 1097 for Public distance ranges. Using this division, we plotted the variation in IoU@0.25 and IoU@0.5 across the distances in Figure \ref{fig:dist_vs_acc}. Across all models with gestural reasoning, we consistently observed a decline in accuracy metrics as distance increased, indicating that 3D-ERU becomes progressively more challenging as the human-to-object distance grows. We also find that the drop in IoU@0.5 is more pronounced for larger distances. In contrast, M3DRefCLIP, a model without gestural reasoning, does not exhibit a decreasing pattern of accuracy as distance increases. To ensure that gesture information is completely excluded from this study on M3DRefCLIP, we removed the human avatar from the scene. On the other hand, \names\textsubscript{ConstantLang}, a model completely relying on pointing gestures, exhibits a clear decreasing pattern of accuracy with increasing human-to-object distance.

        

\section{Conclusion}
\label{sec:conclusion}
\vspace{-0.1in}
We have demonstrated the design and performance of an enhanced model for 3D-ERU tasks, where humans refer to target objects via both gesture and verbal cues. We first introduced the \aug framework to automatically augment existing 3D-REC datasets with pointing gesture cues and used it to create a challenging benchmark dataset for 3D-ERU, named \dataset. We then introduced the novel \namess model that jointly (a)~localizes the human in the 3D scene, (b)~comprehends the pointing gesture, and (c)~fuses language and gestural cues to accurately identify the 3D location of the target object. \namess and \dataset establish a new benchmark for the 3D-ERU task, by achieving 29.5\% higher accuracy compared to prior work.


\section{Acknowledgments}
\vspace{-0.1in}This work was supported in part by: 
    1)  National Research Foundation, Prime Minister’s Office, Singapore under its Campus for Research Excellence and Technological Enterprise (CREATE) program. The Mens, Manus, and Machina (M3S) is an interdisciplinary research group (IRG) of the Singapore-MIT Alliance for Research and Technology (SMART) centre; 
    2) BITS Pilani's Grant No. GOA/ACG/2021-2022/Nov/05; 
    3) Science \& Engineering Research Board's SERB-SURE project number SUR/2022/002735; 
    4) Agency for Science, Technology and Research, Singapore under Grant \#A18A2b0046. 
Any opinions, findings and conclusions or recommendations expressed in this material are those of the author(s) and do not reflect the views of the funding organizations.

{
    \small
    \bibliographystyle{ieeenat_fullname}
    \bibliography{main}
}


\end{document}